\pdfoutput=1
\documentclass[10pt,twocolumn,letterpaper]{article}
\usepackage[textwidth=6.9in,textheight=9.125in,top=0.85in,columnsep=0.3in]{geometry}
\usepackage[T1]{fontenc}
\usepackage{times}
\usepackage{graphicx}
\usepackage[dvipsnames]{xcolor}
\usepackage{amsmath,amssymb}
\usepackage{booktabs,tabularx,multirow,makecell}
\usepackage[numbers,sort&compress]{natbib}
\usepackage[font=small,labelfont=bf,labelsep=period]{caption}
\usepackage{algorithm,algorithmic}
\usepackage{enumitem}
\setlist[itemize]{noitemsep,leftmargin=*,topsep=0.3em}
\usepackage[hyphens]{url}
\definecolor{linkblue}{rgb}{0.21,0.49,0.74}
\usepackage[breaklinks,colorlinks,allcolors=linkblue]{hyperref}
\hypersetup{pdfauthor={Yifeng Lin, Liuxiang Qiu, Yunfeng Chen, Tiesong Zhao},pdftitle={Perception Is Enough, at the Right Scale: Scale-Adaptive Token Pruning for Vision-Language-Action Models}}

\newcommand{\VLAVenue}[1]{%
  \hspace{2pt}{\footnotesize\mdseries\color{black!65}[#1]}%
}
\title{Perception Is Enough, at the Right Scale: \\Scale-Adaptive Token Pruning for Vision-Language-Action Models}
\author{Yifeng Lin$^1$ \quad Liuxiang Qiu$^1$ \quad Yunfeng Chen$^1$ \quad Tiesong Zhao$^{1,2}$\thanks{Corresponding author.}\\[0.5em]
\parbox[t]{0.96\textwidth}{\centering\small
$^1$Fujian Key Laboratory of Intelligent Processing and Wireless Transmission of Media Information, College of Physics and Information Engineering, Fuzhou University, Fuzhou 350108, China.\\[0.5em]
$^2$MoE Key Laboratory of Brain-inspired Intelligent Perception and Cognition, University of Science and Technology of China, Hefei 230027, China.}}
\date{}

\begin{document}
\maketitle

\begin{abstract}
Vision-language-action (VLA) models must process continuous streams of visual observations, and the resulting computational cost limits real-time robotic control. Token pruning reduces this cost, but a fixed semantic scale of perceptual representation can overlook information needed at different stages of manipulation. We show that perceptual semantics can guide effective token selection without relying on internal attention when the selection scale adapts to the scene's evolving semantic structure. Based on this observation, we introduce TauPrune-VLA, a training-free method that controls the scale of the retained perceptual representation through a single perception temperature $\tau$. To adapt this scale as manipulation proceeds, TauPrune-VLA updates $\tau$ online using trends in perceptual semantics, with robot-state changes and action magnitude regulating the adjustment strength. While these semantics evolve over the course of manipulation, they remain relatively stable between consecutive frames. This continuity allows retention indices computed from the current observation to be applied at the next policy invocation, enabling token selection to run asynchronously with inference and reducing pruning overhead. Evaluated across diverse VLA models and simulation benchmarks, TauPrune-VLA achieves up to a $1.86\times$ inference speedup while maintaining near-baseline task success rates. We further validate its effectiveness on two real-world robotic platforms. \href{https://lin0e0.github.io/TauPrune-VLA/}{Project page.}
\end{abstract}

\section{Introduction}
\label{sec:introduction}

\begin{figure}[t]
    \centering
    \includegraphics[width=0.9\columnwidth]{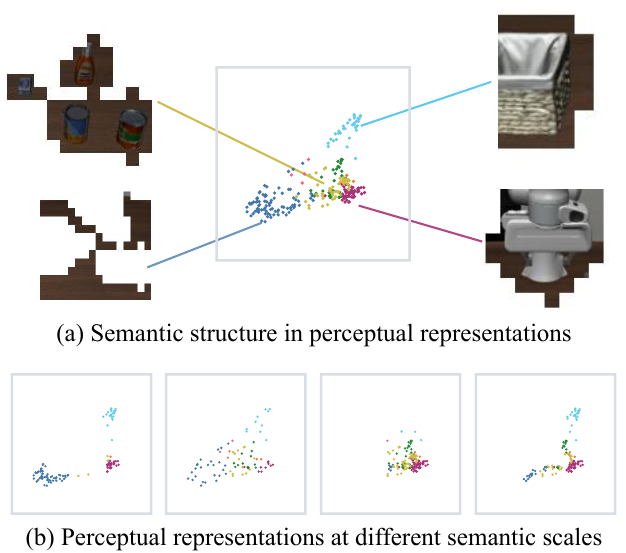}
    \caption{Under a fixed token budget, selecting the semantic scale of perceptual representations entails a trade-off between within-cluster fidelity and cross-cluster coverage.}
    \label{fig:f1}
\end{figure}

Vision-language-action (VLA) models are powerful architectures for embodied AI~\cite{rt2,openvla}. They leverage pretrained vision-language models (VLMs)~\cite{llava,karamcheti2024prismatic} to ground manipulation decisions in semantic information. Real-time deployment requires VLA models to process continuous streams of visual observations, incurring substantial computational overhead. A growing body of work addresses this cost by reducing redundant computation in the VLM backbone.

Common approaches to accelerating VLA inference include architecture-based~\cite{liu2024robomamba}, caching-based~\cite{xu2026vla}, and pruning-based methods~\cite{yang2025efficientvla,li2026spvla}. Token pruning is particularly promising because it is straightforward to implement and requires no architectural modifications. A central question is which tokens to retain. VLA-Pruner~\cite{liu2025vla} combines semantic and action importance scores and uses the combined scores to filter tokens. SpecPrune-VLA~\cite{wang2026specprune} reuses global attention from the previous inference step and applies further dynamic pruning across layers. SAFE-Pruner~\cite{ma2026safe} predicts deep-layer semantic saliency to avoid prematurely discarding tokens needed by deeper layers. Other studies address complementary aspects of pruning, including pruning aggressiveness in VLA-ADP~\cite{pei2026action} and structured sparsity in EcoVLA~\cite{huang2026environment}. Despite their differing emphases, most of these methods rely primarily on internal attention to select tokens, motivated by concerns that perception-based selection may discard action-critical details.

 We argue that this failure comes from the scale at which perceptual signals are read, not from the signals themselves. The visual features of a manipulation scene form semantic structures of very different sizes: a few large structures span most of the tokens, and many small structures span only a few. As illustrated in Fig.~\ref{fig:f1}, token selection at a coarse scale favors large structures and can overlook smaller ones. At a fine scale, the retained tokens capture smaller structures but may represent larger ones less faithfully.  The scale of perceptual representation required for decision-making also changes from one decision step to the next as the robot moves and the semantic structure of its observations evolves. A rule fixed at one scale therefore discards tokens that some steps require. Read at a scale that follows these changes, perceptual semantics can guide VLA token pruning directly. Accordingly, we propose TauPrune-VLA, a token pruning method that adapts the scale of the retained perceptual representation.

TauPrune-VLA uses a perception temperature $\tau$ to adapt token selection to scene-dependent variations in semantic scale. To track perceptual changes over successive decisions, TauPrune-VLA uses trends in perceptual semantics to determine the direction of each temperature update and modulates the step size based on changes in robot state and the magnitude of the robot's actions. TauPrune-VLA also exploits the temporal stability of perceptual semantics to prune asynchronously. It computes token retention indices from the current frame and applies them at the next policy invocation. Our contributions are as follows.
\begin{itemize}
    \item We propose TauPrune-VLA, a token pruning method that accelerates VLA inference by adapting token selection to changing demands for perceptual representations across scales in real-time robotic manipulation.
    \item We show that the effectiveness of perception-based token pruning for VLAs depends on whether perceptual representations are interpreted at a scale suited to the scene's semantic structure at each decision-making step.
    \item We demonstrate that TauPrune-VLA substantially accelerates inference while largely preserving task performance across VLA models and robotic platforms.
\end{itemize}

\section{Related Work}
\label{sec:related_work}

\textbf{Vision-Language-Action Models.}
Vision-language-action (VLA) models extend pretrained vision-language models (VLMs) with action prediction, transferring visual-semantic knowledge to embodied control~\cite{brohan2022rt,rt2,openvla,kim2025fine}. For example, RT-1~\cite{brohan2022rt} learns scalable language-conditioned policies from large-scale robot demonstrations, while RT-2~\cite{rt2} transfers pretrained VLM knowledge to robotic control by representing actions as text tokens. OpenVLA~\cite{openvla} and $\pi_{0.5}$ ~\cite{intelligence2025pi_} adapt pretrained VLM backbones using large-scale, heterogeneous robot data to enable robust visuomotor control. Despite their growing capabilities, VLA models still incur high inference latency, largely due to the cost of processing dense visual tokens in large VLM backbones~\cite{touvron2023llama,team2024gemma} (usually over 2B parameters). This latency limits policy update rates and responsiveness to environmental changes, hindering closed-loop control and efficient task execution.

\noindent\textbf{Efficient Vision-Language-Action Models.} Existing VLA acceleration methods include architecture-based~\cite{liu2024robomamba}, caching-based~\cite{xu2026vla}, and pruning-based approaches~\cite{yang2025efficientvla,li2026spvla}. Architectural methods streamline computation through state-space backbones (e.g., RoboMamba \cite{liu2024robomamba}), compact VLMs (e.g., SmolVLA~\cite{shukor2025smolvla} and EVO-1~\cite{lin2026evo}), or single-pass action heads (e.g., EffVLA~\cite{sun2026makes}). Caching-based methods exploit temporal redundancy, such as VLA-Cache~\cite{xu2026vla} reuses cached KV representations of temporally stable visual tokens while recomputing task-relevant ones, whereas LAC~\cite{wei2026learning} learns token selection and cache ratios.

Pruning offers a promising complementary approach by shortening visual sequences through attention-based, perception-based, or hybrid token selection. A key challenge is retaining task-relevant visual evidence under limited token budgets. Attention-based methods exploit internal attention signals across timesteps to estimate token saliency and shorten visual sequences~\cite{wang2026specprune,ma2026safe}. Perception-based methods use the structure of visual features to guide token pruning in VLMs. For instance, DivPrune~\cite{alvar2025divprune} captures the diversity among visual tokens for effective pruning, and CDPruner~\cite{zhang2026beyond} jointly preserves conditional diversity and instruction relevance to perform token pruning. Hybrid methods incorporate perceptual cues but remain largely attention-driven. EfficientVLA~\cite{yang2025efficientvla} supplements attention-selected tokens with feature diversity, while SP-VLA~\cite{li2026spvla} combines semantic attention with object contours. Most methods favor attention over perception to avoid losing action-critical details. We attribute this risk to perceptual scale mismatch and propose TauPrune-VLA, a scale-adaptive token pruning framework for VLA models.

\section{Methodology}
\label{sec:methodology}

\begin{figure*}[t]
    \centering
    \includegraphics[width=\textwidth]{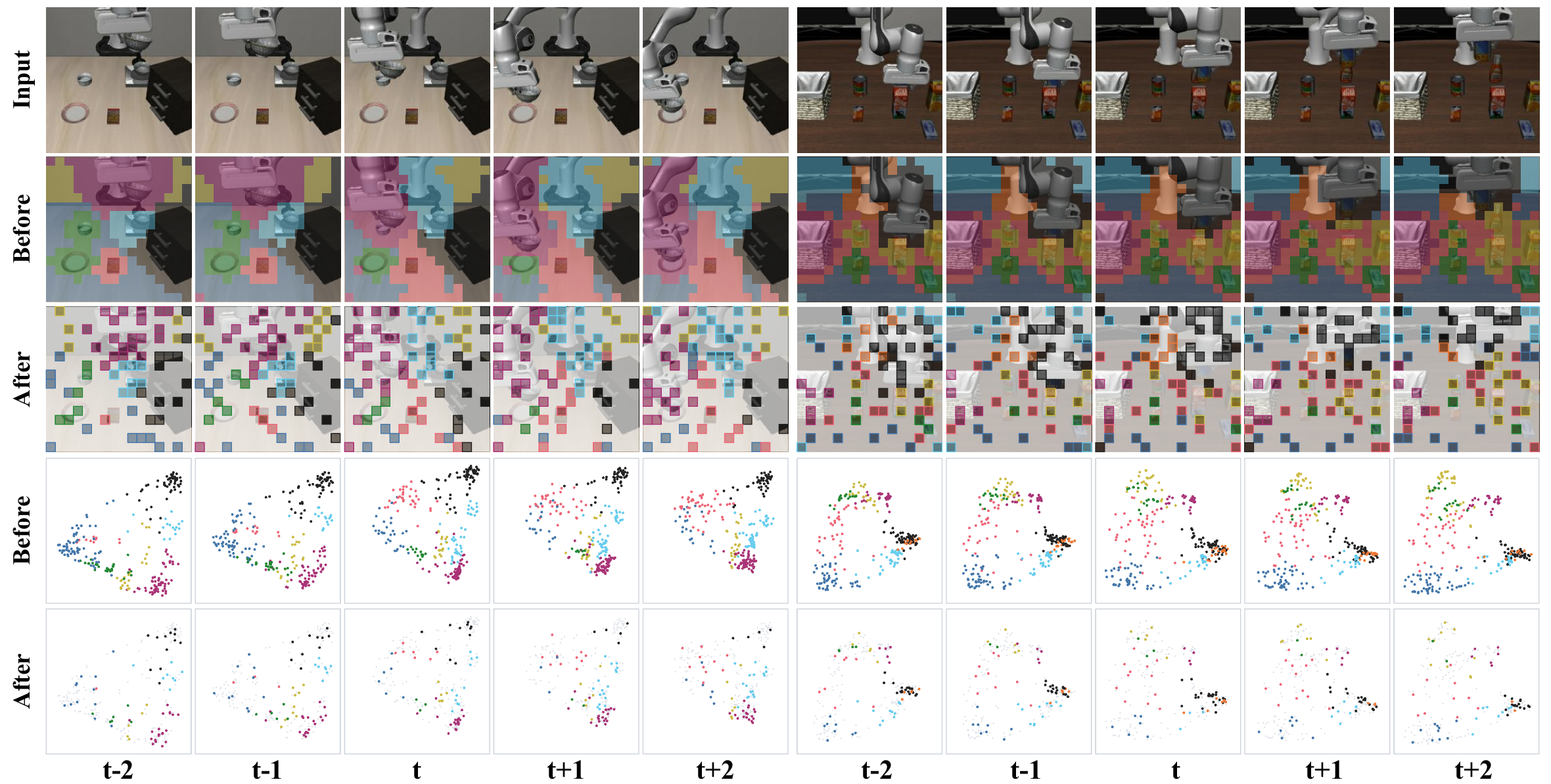}
    \caption{Feature clustering and PCA projections before and after TauPrune-VLA pruning on LIBERO~\cite{liu2023libero}.}
    \label{fig:yuyi}
\end{figure*}

\subsection{Problem Formulation}
\label{sec:Problem Formulation}

Given a sequence of tokens $X=\{x_1,x_2,\ldots,x_N\}$ at the pruning point and a retention budget $K<N$, VLA token pruning selects an index subset $I\subseteq\{1,\ldots,N\}$ to form the compressed sequence: $S=X[I],|I|=K$. The retained tokens preserve their original order and position indices. Both inputs use the same sequence $C$ of tokens from non-visual sources. Let $g_\theta$ denote downstream mapping that performs decoding and predicts actions. The action predictions for the full and pruned inputs are $a=g_\theta(X,C)$, and $a_I=g_\theta(S,C)$.
The resulting budget-constrained objective is
\begin{equation}
\label{eq:pruning-objective}
\min_{I\subseteq\{1,\ldots,N\}}
\mathcal{D}\!\left(a,\,a_I\right)
\quad
\text{s.t.}\quad |I|=K,
\end{equation}
where $\mathcal{D}$ measures the discrepancy between action predictions. We hold model parameters $\theta$ fixed, and $K$ limits the number of tokens processed downstream. From a perceptual perspective, when the token distribution of $S$ approximates that of $X$, we expect the mapping $g_\theta$ to yield a small discrepancy $\mathcal{D}$ under identical computational constraints.

\begin{figure}[t]
    \centering
    \includegraphics[width=\columnwidth]{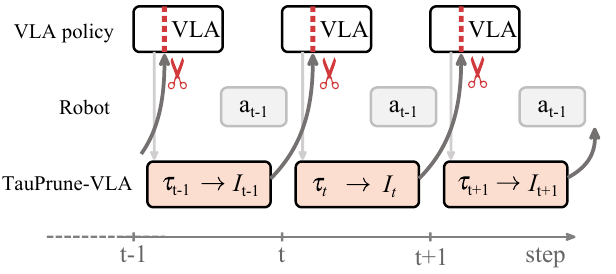}
    \caption{Asynchronous pruning architecture of TauPrune-VLA with pruning decisions computed in advance.}
    \label{fig:f2}
\end{figure}

\subsection{Semantic Scale of Perceptual Representations}
\label{sec:Semantic Scale of Perceptual Representations}

Clustering results and PCA projections of perceptual features from OpenVLA-OFT on LIBERO suggest that semantic structure is relatively stable across adjacent observation frames (Fig.~\ref{fig:yuyi}). The observed changes mainly involve shifts and rescaling of semantic clusters. Here, changes in semantic scale refer broadly to these variations in cluster positions and extents. Within a single frame, semantic clusters vary in size and separation. A fixed rule for selecting tokens from perceptual signals can therefore struggle to balance coverage across clusters with fidelity within clusters. Even if a selection scale is optimal for the current frame, it may become unsuitable as the robot executes successive actions and the perceptual representations evolve. These changes also exhibit temporal regularity. As illustrated in Fig.~\ref{fig:yuyi}, semantic structures typically contract or expand gradually across decision steps. Larger state changes or action magnitudes are associated with larger changes in semantic scale along the prevailing trend. These observations motivate TauPrune-VLA, a scale-adaptive token pruning method for vision-language-action models. By adapting to changes in semantic scale, TauPrune-VLA aims to reduce the discrepancy between perceptual representations before and after pruning, as illustrated in Fig.~\ref{fig:yuyi}. This representation preservation provides a perceptual strategy for pursuing the pruning objective $\min_{I\subseteq\{1,\ldots,N\}}
\mathcal{D}\!\left(a,\,a_I\right)$.

\subsection{Pipelined Token Pruning}
\label{sec:Cross-Layer Index Mapping and Asynchronous Pruning}

As discussed in Section~\ref{sec:Semantic Scale of Perceptual Representations}, perceptual structure remains relatively stable across adjacent frames, with most changes involving the rescaling of semantic clusters.  Based on this observation, TauPrune-VLA integrates token selection into VLA inference (Fig.~\ref{fig:f2}). It computes and reuses retention indices asynchronously across policy invocations and applies them to prune tokens at a predetermined layer.

At step $t$, we apply Perceptual Token Selection to visual features $X_t^v$ to obtain the retention indices $I_t$. We use these indices to prune the token sequence $X_{t+1}$ at step $t+1$:
\begin{equation}
S_{t+1}=X_{t+1}[I_t],
\end{equation}
where $X_{t+1}$ denotes the token sequence at the VLA model layer. The pruning layer can be selected at different depths. By default, TauPrune-VLA prunes tokens after the second decoder layer. The temperature $\tau$ used for Perceptual Token Selection at step $t$ is computed during the execution of the decision at step $t-1$. Together, these stages form an asynchronous pipeline whose stages operate at different rates.

\subsection{Perceptual Token Selection}
\label{sec:Perceptual Token Selection}

Given a nonempty retained set $S\subseteq X$, the representation residual of token $i$ is
\begin{equation}
\label{eq:representation-residual}
r_i(S)=\min_{j\in I}d_{ij},
\end{equation}
where $d_{ij}$ is the chord distance between tokens $i$ and $j$. A smaller residual indicates $S$ better represents token $i$. When $S$ is empty, we set $r_i(\varnothing)=2$ for every token, using the upper bound on the chord distance. The local representation gain $\delta_i(j\mid S)$ for token $i$ from adding candidate $j$ is
\begin{equation}
\label{eq:local-representation-gain}
\begin{aligned}
\delta_i(j\mid S)
&=r_i(S)-r_i(S\cup\{j\})\\
&=r_i(S)-\min\{r_i(S),d_{ij}\}\\
&=[r_i(S)-d_{ij}]_+,
\end{aligned}
\end{equation}
where $[a]_+=\max\{a,0\}$. The gain is positive when $d_{ij}<r_i(S)$ and zero otherwise. Thus, the value of a candidate depends on how much it improves the representation provided by the current retained set. Averaging the local gains gives the global representation gain $G(j\mid S)$:
\begin{equation}
\label{eq:global-representation-gain}
G(j\mid S)=\frac{1}{N}\sum_{i=1}^{N}\delta_i(j\mid S).
\end{equation}
This gain serves as an proxy for how adding token $j$ improves representation of the perceptual semantic distribution.

In VLA decision-making, scene understanding, object understanding, and fine-grained manipulation place different demands on perceptual semantics. The corresponding local gains can contribute to the overall improvement at different scales. As shown in Fig.~\ref{fig:yuyi}, semantic clusters differ in their spread, and the distances between them vary across cluster pairs. We therefore introduce a scaling factor $w_i$ for each local gain to adjust the global score. For a fixed candidate $j$ and retained set $S$, write $\delta_i=\delta_i(j\mid S)$. When at least one local gain is positive, we define
\begin{equation}
\label{eq:weighted-gain-definition}
\begin{aligned}
G_\kappa(j\mid S)
&=\frac{1}{N}\sum_{i=1}^{N}w_i\,\delta_i(j\mid S),\\
w_i
&=\left(
    \frac{\delta_i(j\mid S)}{G_\kappa(j\mid S)}
  \right)^\kappa.
\end{aligned}
\end{equation}
where $\kappa>0$ controls the weighting. Setting perception temperature $\tau=1/(\kappa+1)$ and solving for $G(j\mid S)$ yields the adjusted global representation gain $G_\tau(j\mid S)$:
\begin{equation}
\label{eq:adjusted-global-gain}
G_\tau(j\mid S)
=
\left[
    \frac{1}{N}\sum_{i=1}^{N}
    \left([r_i(S)-d_{ij}]_+\right)^{1/\tau}
\right]^\tau.
\end{equation}

At each iteration, we score unselected indices using
$G_\tau(j\mid S)$, add the highest-scoring index to $I$, and update
the representation residuals. We repeat until $|I|=K$.
Algorithm~\ref{alg:greedy_token_selection} summarizes the procedure
and returns the selected indices in ascending order. The greedy
updates reduce the representation gap between the
retained subset and the full token sequence. By adapting the
perception temperature $\tau$, TauPrune-VLA adjusts the selection
criterion to the perceptual scale required at each decision step.

\begin{algorithm}[t]
\caption{Perceptual Token Selection}
\label{alg:greedy_token_selection}
\small
\begin{algorithmic}[1]
\REQUIRE Nonzero visual features $X^v=\{x_i^v\}_{i=1}^{N}$;
         retention budget $1\leq K<N$;
         perception temperature $\tau\in(0,1]$.
\ENSURE $K$ distinct retention indices, sorted in ascending order.

\STATE $\widehat{x}_i^v \gets
        x_i^v/\lVert x_i^v\rVert_2,
        \quad i=1,\ldots,N$
\STATE $d_{ij} \gets
        \lVert\widehat{x}_i^v-\widehat{x}_j^v\rVert_2,
        \quad i,j=1,\ldots,N$
\STATE $I \gets \varnothing$
\STATE $r_i \gets 2,\quad i=1,\ldots,N$

\FOR{$k=1,\ldots,K$}
    \FORALL{$j\in\{1,\ldots,N\}\setminus I$}
        \STATE $\delta_i(j\mid S) \gets
                [r_i-d_{ij}]_+,
                \quad i=1,\ldots,N$
        \STATE $G_\tau(j\mid S) \gets
                \left[
                    \frac{1}{N}\sum_{i=1}^{N}
                    \delta_i(j\mid S)^{1/\tau}
                \right]^\tau$
    \ENDFOR

    \STATE $j^\star \gets
        \min\operatorname*{arg\,max}_{j\in\{1,\ldots,N\}\setminus I}
        G_\tau(j\mid S)$
    \STATE $I \gets I\cup\{j^\star\}$
    \STATE $r_i \gets \min\{r_i,d_{i j^\star}\},
            \quad i=1,\ldots,N$
\ENDFOR

\RETURN $\operatorname{SortAscending}(I)$
\end{algorithmic}
\end{algorithm}

\begin{figure*}[t]
    \centering
    \includegraphics[width=\textwidth]{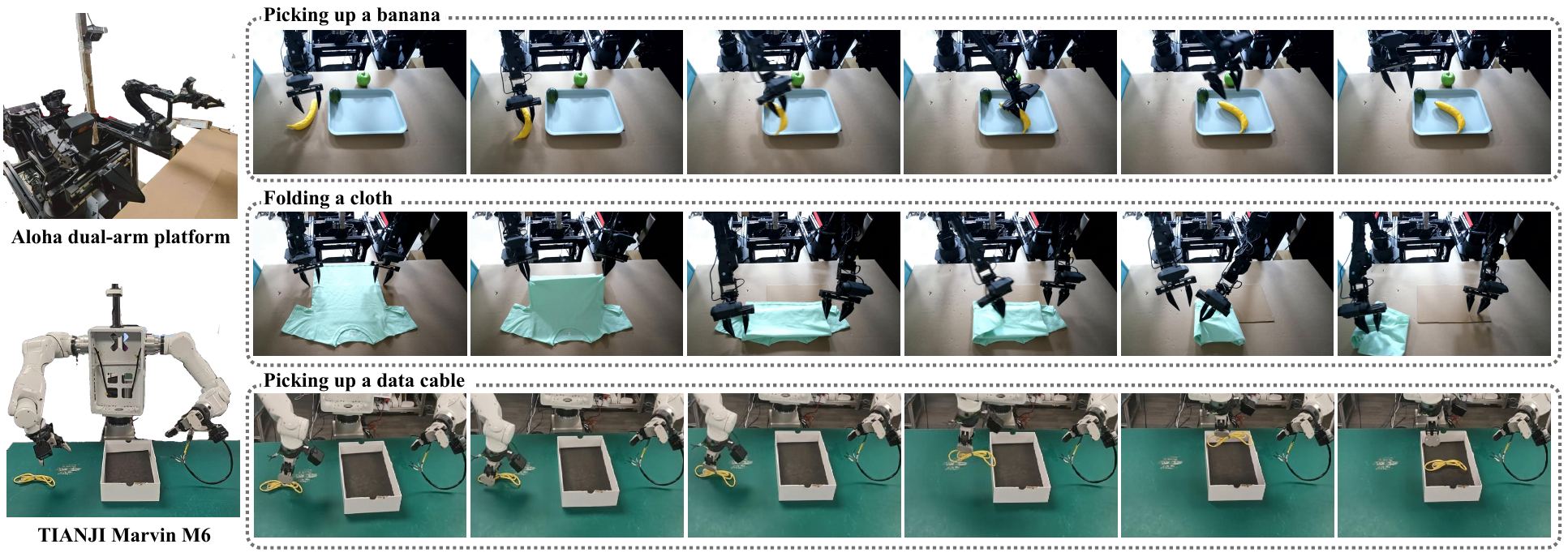}
    \caption{\textbf{Real-robot platforms and task execution.} \textbf{Left:} The Aloha dual-arm platform and the TIANJI Marvin M6, both equipped with 7-DoF robotic arms, grippers, and wrist- and head-mounted cameras. \textbf{Right:} Task execution sequences of $\pi_{0.5}$ with TauPrune-VLA, illustrating continuous execution on the physical robots.}
    \label{fig:real_robot_setup}
\end{figure*}

\subsection{Perception--Action Temperature Modulation}

During robotic manipulation, perceptual semantics evolve as the robot moves and interacts with the environment, changing the distribution of local representation gains. As discussed in Section~\ref{sec:Semantic Scale of Perceptual Representations}, semantic clusters may shift or change scale when action magnitudes vary substantially. TauPrune-VLA therefore adjusts the temperature according to changes in perceptual semantic structure and uses recent motion intensity to modulate the adjustment magnitude. This allows the adjusted global representation gain to vary as manipulation proceeds. At the $t$-th policy invocation, we construct a Gaussian kernel matrix from $N$ visual features of the primary view and normalize it by its trace. Let $\{\lambda_{t,i}\}_{i=1}^{N}$ denote the eigenvalues of the resulting matrix. We compute the kernel spectral entropy $H_t$ and kernel purity $P_t$ to measure the dispersion and concentration of the kernel spectrum, respectively:
\begin{equation}
\label{eq:kernel-entropy-purity}
H_t=-\sum_{i=1}^{N}\lambda_{t,i}\log\lambda_{t,i},
\qquad
P_t=\sum_{i=1}^{N}\lambda_{t,i}^{2}.
\end{equation}
We use complementary statistics to characterize perceptual structure. The normalized change in spectral entropy $\Delta\bar H_t$ and the change in kernel purity $\Delta P_t$ serve as proxy signals for the evolution of perceptual semantic distribution,
\begin{equation}
\label{eq:kernel-statistic-changes}
\Delta\bar H_t=\frac{H_t-H_{t-1}}{\log N},
\qquad
\Delta P_t=P_t-P_{t-1}.
\end{equation}

To reduce sensitivity to fluctuations, we assign a nonzero direction signal $u_t$ when both statistics support the same trend and their changes exceed corresponding thresholds:
\begin{equation}
\label{eq:temperature-direction-signal}
u_t=
\begin{cases}
1, & \Delta\bar H_t>\epsilon_H
      \ \text{,}\ \Delta P_t<-\epsilon_P,\\[2pt]
-1, & \Delta\bar H_t<-\epsilon_H
      \ \text{,}\ \Delta P_t>\epsilon_P,\\[2pt]
0,  & \text{otherwise}.
\end{cases}
\end{equation}
We set both change thresholds, $\epsilon_H$ and $\epsilon_P$, to $2.5\times10^{-4}$ in all experiments. A signal of $u_t=1$ indicates that the perceptual structure is becoming more dispersed and calls for a higher target temperature, placing more emphasis on gains accumulated across multiple locations. A signal of $u_t=-1$ indicates that the structure is becoming more concentrated and calls for a lower target temperature, increasing the relative contribution of larger local gains. Small changes or disagreement between the two statistics yield $u_t=0$.

We use recent motion intensity $m_{t-1}$ to modulate effect of perceptual feedback on temperature. It is computed from normalized past states $\mathbf{s}_{t-1}$, $\mathbf{s}_{t-2}$ and the latest action $\mathbf{a}_{t-1}$,
\begin{equation}
\label{eq:recent-motion-intensity}
m_{t-1}
=
\operatorname{clip}_{[0,1]}\!\left(
    \max\left\{
        \lVert\mathbf{s}_{t-1}-\mathbf{s}_{t-2}\rVert_\infty,
        \lVert\mathbf{a}_{t-1}\rVert_\infty
    \right\}
\right).
\end{equation}

We initialize the temperature as $\tau_0=\tau_{\mathrm{base}}$. At each subsequent invocation, we combine perceptual structure feedback with recent motion to compute the target temperature,
\begin{equation}
\label{eq:target-temperature}
\widehat{\tau}_t
=
\operatorname{clip}_{[\tau_{\min},\tau_{\max}]}\!\left(
    \tau_{\mathrm{base}}+\beta m_{t-1}u_t
\right).
\end{equation}
We then apply exponential smoothing by coefficien $\alpha$,
\begin{equation}
\label{eq:smoothed-temperature}
\tau_t=(1-\alpha)\tau_{t-1}+\alpha\widehat{\tau}_t.
\end{equation}
This formulation separates the direction of temperature adaptation from its magnitude: perceptual feedback determines the direction of the target's deviation from $\tau_{\mathrm{base}}$, while recent motion scales its magnitude. The baseline temperature $\tau_{\mathrm{base}}$ provides a stable reference for modulation: when $u_t=0$ or $m_{t-1}=0$, the target returns to the baseline, and the temperature relaxes toward it. For a fixed target and $0<\alpha<1$, the deviation from that target contracts by a factor of $1-\alpha$ at each invocation, yielding geometric convergence. Together with clipping to $[\tau_{\min},\tau_{\max}]$ with $\tau_{\min}>0$, this update prevents temperature collapse and unbounded growth. We empirically set the smoothing coefficient to $\alpha=0.5$ and the baseline temperature to $\tau_{\mathrm{base}}=0.25$ for all experiments. The nonnegative modulation coefficient $\beta$ is set to either $0.25$ or $0.75$. Unless otherwise specified, the same hyperparameter settings are used for the head-mounted and wrist-mounted cameras. 

\begin{table*}[t]
  \centering
  \caption{\textbf{Comparison of inference acceleration methods on LIBERO
  for OpenVLA-OFT and $\boldsymbol{\pi_{0.5}}$.}
  Average denotes the mean success rate across four task suites.
  Speedup is relative to each model's Vanilla baseline.
  \textbf{Bold} marks the best value in each column, including ties.}
  \label{tab:vla-accuracy-efficiency}
  \small
  \setlength{\tabcolsep}{1.15pt}
  \setlength{\aboverulesep}{0pt}
  \setlength{\belowrulesep}{0pt}
  \renewcommand{\arraystretch}{1.23}
  \newcommand{\VLAHeading}[1]{{\footnotesize\mdseries #1}}
  \newcommand{\VLAUnit}[1]{{\footnotesize #1}}
  \begin{tabular*}{\linewidth}{@{}l@{\hspace{2pt}}|
      c@{\extracolsep{\fill}\hspace{2\tabcolsep}}*{4}{c}|crc
      @{\extracolsep{0pt}\hspace{3.8pt}}|
      c@{\extracolsep{\fill}\hspace{2\tabcolsep}}*{4}{c}|crc@{}}
    \toprule
    & \multicolumn{8}{c@{\hspace{3.8pt}}|}{OpenVLA-OFT}
    & \multicolumn{8}{c@{}}{$\pi_{0.5}$} \\
    \cline{2-17}
    \VLAHeading{Method}
    & \multicolumn{5}{c|}{\VLAHeading{Success Rate}~\VLAUnit{(\%)~$\uparrow$}}
    & \VLAHeading{FLOPs} & \multicolumn{1}{c}{\VLAHeading{Latency}} & \VLAHeading{Speedup}
    & \multicolumn{5}{c|}{\VLAHeading{Success Rate}~\VLAUnit{(\%)~$\uparrow$}}
    & \VLAHeading{FLOPs} & \multicolumn{1}{c}{\VLAHeading{Latency}} & \VLAHeading{Speedup} \\
    \cline{2-6}\cline{10-14}
    & \VLAHeading{Spatial} & \VLAHeading{Object} & \VLAHeading{Goal}
    & \VLAHeading{Long} & \VLAHeading{Average} & \VLAUnit{(T)~$\downarrow$}
    & \multicolumn{1}{c}{\VLAUnit{(ms)~$\downarrow$}} & \VLAUnit{$\uparrow$}
    & \VLAHeading{Spatial} & \VLAHeading{Object} & \VLAHeading{Goal}
    & \VLAHeading{Long} & \VLAHeading{Average} & \VLAUnit{(T)~$\downarrow$}
    & \multicolumn{1}{c}{\VLAUnit{(ms)~$\downarrow$}} & \VLAUnit{$\uparrow$} \\
    \midrule
    Vanilla
    & 98.2 & 97.8 & 96.6 & 94.8 & 96.85 & 3.97 & 122.57 & 1.00$\times$
    & \textbf{98.8} & 98.0 & 98.2 & \textbf{93.8} & \textbf{97.20} & 1.99 & 67.04 & 1.00$\times$ \\
    FastV\VLAVenue{ECCV24}
    & 98.6 & 97.6 & 95.2 & 89.6 & 95.25 & 1.56 & \textbf{62.92} & \textbf{1.95}$\times$
    & 81.8 & 95.2 & 91.4 & 74.4 & 85.70 & 1.27 & \textbf{37.31} & \textbf{1.80}$\times$ \\
    VLA-Cache\VLAVenue{NIPS25}
    & 98.4 & 96.8 & 96.2 & \textbf{95.0} & 96.60 & 3.09 & 114.55 & 1.07$\times$
    & 98.2 & 98.0 & \textbf{98.8} & 90.0 & 96.25 & 1.87 & 61.41 & 1.09$\times$ \\
    VLA-Pruner\VLAVenue{arXiv25}
    & 98.4 & 98.2 & 96.4 & 92.2 & 96.30 & 2.12 & 82.28 & 1.49$\times$
    & 97.6 & \textbf{98.6} & 98.2 & 91.4 & 96.45 & 1.41 & 50.19 & 1.34$\times$ \\
    DivPrune\VLAVenue{CVPR25}
    & 98.4 & 97.0 & 97.2 & 88.2 & 95.20 & 1.56 & 69.34 & 1.77$\times$
    & 96.2 & 98.4 & 96.8 & 90.0 & 95.35 & 1.27 & 44.92 & 1.49$\times$ \\
    MMTok\VLAVenue{ICLR26}
    & 97.8 & 97.6 & 95.4 & 93.0 & 95.95 & 1.56 & 73.49 & 1.67$\times$
    & 96.6 & 97.2 & 97.4 & 92.4 & 95.90 & 1.27 & 48.34 & 1.39$\times$ \\
    VLA-ADP\VLAVenue{ICLR26}
    & 97.0 & 96.6 & 95.6 & 85.2 & 93.60 & 2.87 & 93.11 & 1.32$\times$
    & 98.2 & 97.6 & 97.4 & 88.8 & 95.50 & 1.64 & 50.69 & 1.32$\times$ \\
    SAFE-Pruner\VLAVenue{ECCV26}
    & \textbf{98.8} & \textbf{98.8} & 96.4 & 90.4 & 96.10 & 1.84 & 96.01 & 1.28$\times$
    & 92.0 & 98.0 & 95.0 & 85.0 & 92.50 & 1.50 & 54.48 & 1.23$\times$ \\
    SpecPrune-VLA\VLAVenue{ICML26}
    & 98.6 & 97.8 & 97.6 & 91.0 & 96.25 & \textbf{1.51} & 86.26 & 1.42$\times$
    & 97.4 & 94.6 & 95.8 & 90.0 & 94.45 & \textbf{1.19} & 55.42 & 1.21$\times$ \\
    \midrule
    \textbf{TauPrune-VLA}
    & 98.2 & 97.6 & \textbf{98.0} & 94.6 & \textbf{97.10} & 1.56 & 65.82 & 1.86$\times$
    & 97.6 & 98.4 & \textbf{98.8} & 92.0 & 96.70 & 1.27 & 38.87 & 1.72$\times$ \\
    \bottomrule
  \end{tabular*}
\end{table*}

\section{Experiments}
\label{sec:experiments}

\subsection{Experimental Settings}
\label{sec:experimental_setup}

\textbf{Baselines.}
We evaluate TauPrune-VLA across open source VLA models, including OpenVLA-OFT~\cite{kim2025fine}, $\pi_{0.5}$~\cite{intelligence2025pi_}, and CogACT~\cite{li2024cogact}.
The unaccelerated version of each model serves as the primary baseline.
We use VLA-Cache~\cite{xu2026vla} as the KV caching baseline.
For token pruning, we include methods developed for VLAs (VLA-ADP~\cite{pei2026action}, SAFE-Pruner~\cite{ma2026safe}, SpecPrune-VLA~\cite{wang2026specprune}, and VLA-Pruner~\cite{liu2025vla}) and VLMs (MMTok~\cite{dong2026mmtok}, DivPrune~\cite{alvar2025divprune}, and FastV~\cite{chen2024image}).

\textbf{Evaluation Protocol.}
Unless otherwise specified, we retain default pruning ratios for open-source methods and set the pruning ratio to \(75\%\) for all other methods, including TauPrune-VLA. For perception-based methods designed to prune VLM tokens, pruning is performed at the second VLM layer to avoid prematurely removing critical tokens. Following the evaluation protocol of VLA-Pruner, we report task success rate (\%), computational cost (TFLOPs), and inference latency (ms). All experiments use an NVIDIA A40 GPU.

\subsection{Evaluation Benchmarks}
\label{sec:evaluation_benchmarks}

\textbf{Simulation Experiments. }We use LIBERO~\cite{liu2023libero} and SIMPLER~\cite{li24simpler} as our evaluation suites for comprehensive performance assessment across diverse robotic manipulation scenarios under experimental conditions. LIBERO assesses VLA generalization across four manipulation dimensions: spatial reasoning, object-centric interaction, goal-directed execution, and long-horizon planning. SIMPLER provides Visual Matching and Variant Aggregation as evaluation settings for a Google robot arm. Each setting covers four representative manipulation tasks, namely Pick Coke Can (PickCan), Move Near (MoveNear), Open / Close Drawer (Drawer), and Open Top Drawer and Place Apple (DrawerApple). Together, these suites assess task-level generalization and robustness to visual variations. 

\textbf{Real-Robot Experiments.} We conduct real-world evaluations on the Aloha dual-arm platform \cite{fu2024mobile} and the TIANJI Marvin M6 equipped with an STS3032 Omnipicker, as shown in Fig.~\ref{fig:real_robot_setup}. On the Aloha dual-arm platform, we deploy fine-tuned OpenVLA-OFT and fully trained $\pi_{0.5}$ as baselines for two tasks: Pick Up a Banana (PickBanana) and Fold a Cloth (FoldCloth). On the TIANJI Marvin M6, we deploy fine-tuned $\pi_{0.5}$ as the baseline for Pick Up a Data Cable (PickCable). The action chunk sizes for OpenVLA-OFT and $\pi_{0.5}$ are $25 \times 14$ and $50 \times 14$, respectively.

\subsection{Main Results}
\label{sec:main_results}

\begin{table}[t]
  \centering
  \caption{\textbf{TauPrune-VLA on SIMPLER.}
    VM and VA denote Visual Matching and Variant Aggregation.
    PC, MN, D, and DA denote PickCan, MoveNear, Drawer, and DrawerApple.
    TP denotes TauPrune-VLA. Speedup is relative to Vanilla in each setting.}
  \label{tab:simpler-results}
  \small
  \setlength{\tabcolsep}{1pt}
  \setlength{\aboverulesep}{0pt}
  \setlength{\belowrulesep}{0pt}
  \renewcommand{\arraystretch}{1.23}
  \newcommand{\SimplerHead}[1]{{\footnotesize #1}}

  \begin{tabular*}{\columnwidth}{@{\extracolsep{\fill}}c|l|ccccc|ccc@{}}
    \toprule
    \multirow{2}{*}{\SimplerHead{Set.}}
    & \multirow{2}{*}{\SimplerHead{Method}}
    & \multicolumn{5}{c|}{\SimplerHead{Success Rate (\%) $\uparrow$}}
    & \SimplerHead{FLOPs} & \SimplerHead{Lat.} & \SimplerHead{Speedup} \\
    \cline{3-7}
    & & \SimplerHead{PC} & \SimplerHead{MN} & \SimplerHead{D}
    & \SimplerHead{DA} & \SimplerHead{Avg.}
    & \SimplerHead{(T) $\downarrow$}
    & \SimplerHead{(ms) $\downarrow$}
    & \SimplerHead{$\uparrow$} \\
    \midrule
    \multirow{2}{*}{VM}
    & Vanilla & 92.3 & 83.3 & 75.0 & 49.1 & 74.93
              & 1.81 & 63.89 & $1\times$ \\
    &  TP    & 92.0 & 82.1 & 75.5 & 51.9 & 75.35
              & 1.22 & 43.53 & $1.47\times$ \\
    \midrule
    \multirow{2}{*}{VA}
    & Vanilla & 89.9 & 76.7 & 31.2 & 50.8 & 62.15
              & 1.81 & 64.22 & $1\times$ \\
    &  TP    & 90.2 & 74.2 & 32.0 & 51.3 & 61.92
              & 1.22 & 44.90 & $1.43\times$ \\
    \bottomrule
  \end{tabular*}
\end{table}

\textbf{Results on LIBERO.} To evaluate the effectiveness of TauPrune-VLA, we compare it with eight acceleration methods on OpenVLA-OFT and $\pi_{0.5}$, as shown in Table~\ref{tab:vla-accuracy-efficiency}. On both backbones, TauPrune-VLA achieves the highest average success rate and the second-highest speedup among the evaluated acceleration methods. On OpenVLA-OFT, TauPrune-VLA reduces computational cost from 3.97 to 1.56 TFLOPs and achieves a $1.86\times$ speedup, while exceeding the average success rate of OpenVLA-OFT. This improvement suggests that token pruning that adapts to the scale of perceptual representations can filter out noisy tokens that are unhelpful for decision-making while substantially reducing overall computational redundancy. Similarly, on $\pi_{0.5}$, TauPrune-VLA further reduces computation and achieves a $1.72\times$ speedup. Its average success rate is only 0.50\% lower than that of $\pi_{0.5}$ and remains the highest among the evaluated acceleration methods consistently. 

TauPrune-VLA maintains perceptual representations over time at low computational cost. Its strong performance retention on LIBERO-Long suggests limited temporal accumulation of errors associated with representation residuals. It also achieves faster inference than methods with comparable FLOPs, an advantage we attribute largely to its asynchronous execution architecture. The qualitative results in Fig.~\ref{fig:yuyi} further illustrate its ability to maintain perceptual representations. TauPrune-VLA consistently preserves perceptual semantic structure, helping maintain coverage across semantic clusters and fidelity within each cluster while rapidly adapting to changes in perceptual representations. These observations support Perceptual Token Selection based on perceptual representation scale, together with Perception--Action Temperature Modulation for sustaining its benefits during sequential decision-making.

TauPrune-VLA also maintains strong performance across task attributes, suggesting better retention of the accelerated VLA models' generalization capabilities on the evaluated suites. The suite-level success rates indicate that perception-based pruning methods designed for VLMs, such as MMTok and DivPrune, and methods that use action attention or attention from deeper layers, such as SpecPrune-VLA and SAFE-Pruner, tend to preserve performance on some attributes at the expense of others. By adapting to the scale of perceptual representations, TauPrune-VLA achieves more balanced success rates across suites, particularly on OpenVLA-OFT. These results highlight the importance of exploiting perceptual representations at an appropriate scale for VLA token pruning.

\begin{table}[t]
  \centering
  \caption{TauPrune-VLA on PickBanana and FoldCloth.}
  \label{tab:pickbanana-foldcloth-results}
  \small
  \setlength{\tabcolsep}{2pt}
  \setlength{\aboverulesep}{0pt}
  \setlength{\belowrulesep}{0pt}
  \renewcommand{\arraystretch}{1.18}
  \newcommand{\PickFoldHead}[1]{{\footnotesize #1}}

  \begin{tabularx}{\columnwidth}{
    @{}l|
    *{2}{>{\centering\arraybackslash}X}|
    *{3}{>{\centering\arraybackslash}X}
  }
    \toprule
    \rule[-3pt]{0pt}{21pt}\raisebox{4pt}{\PickFoldHead{Method}}
    & \PickFoldHead{\shortstack{Pick\\Banana}}
    & \PickFoldHead{\shortstack{Fold\\Cloth}}
    & \PickFoldHead{\shortstack{FLOPs\\(T)~$\downarrow$}}
    & \PickFoldHead{\shortstack{Latency\\(ms)~$\downarrow$}}
    & \PickFoldHead{\shortstack{Speedup\\$\uparrow$}} \\
    \midrule
    OpenVLA-OFT    & 28/30 &  6/30 & 7.73 & 183.80 & $1\times$ \\
    + TauPrune-VLA & 29/30 &  5/30 & 3.99 &  110.72 & $1.66\times$ \\
    \midrule
    $\pi_{0.5}$    & 29/30 & 26/30 & 1.99 &  70.31 & $1\times$ \\
    + TauPrune-VLA & 29/30 & 26/30 & 0.92 &  45.70 & $1.54\times$ \\
    \bottomrule
  \end{tabularx}
\end{table}

\begin{table}[t]
  \centering
  \caption{TauPrune-VLA on PickCable.}
  \label{tab:pickcable-results}
  \small
  \setlength{\tabcolsep}{2pt}
  \setlength{\aboverulesep}{0pt}
  \setlength{\belowrulesep}{0pt}
  \renewcommand{\arraystretch}{1.18}
  \newcommand{\PickCableHead}[1]{{\footnotesize #1}}

  \begin{tabularx}{\columnwidth}{
    @{}l|>{\centering\arraybackslash}X|
    *{3}{>{\centering\arraybackslash}X}
  }
    \toprule
    \rule[-3pt]{0pt}{21pt}\raisebox{4pt}{\PickCableHead{Method}}
    & \PickCableHead{\shortstack{Pick\\Cable}}
    & \PickCableHead{\shortstack{FLOPs\\(T)~$\downarrow$}}
    & \PickCableHead{\shortstack{Latency\\(ms)~$\downarrow$}}
    & \PickCableHead{\shortstack{Speedup\\$\uparrow$}} \\
    \midrule
    $\pi_{0.5}$    & 26/30 & 1.99 & 76.60 & $1\times$ \\
    + TauPrune-VLA & 24/30 & 0.92 & 58.69 & $1.31\times$ \\
    \bottomrule
  \end{tabularx}
\end{table}

\textbf{Results on SIMPLER.} We further evaluate TauPrune-VLA on CogACT to assess its applicability to another type of VLA model. Table~\ref{tab:simpler-results} reports results on the SIMPLER benchmark under the VM and VA settings, with pruning performed at the second layer using a $37.5\%$ pruning ratio. Under VM, TauPrune-VLA improves both average success rate and inference speed over the CogACT baseline. Under VA, it accelerates inference with an average success rate only 0.23\% below that of the baseline. These findings provide further evidence that TauPrune-VLA's perceptually guided pruning strategy transfers across VLA models.

\textbf{Results on Real Robot.}
To assess the real-world generalization of TauPrune-VLA, we report results from 30 trials per task on two robotic platforms in Tables~\ref{tab:pickbanana-foldcloth-results} and~\ref{tab:pickcable-results}. Fig.~\ref{fig:real_robot_setup} illustrates the execution of $\pi_{0.5}$ with TauPrune-VLA.

On the Aloha dual-arm platform, TauPrune-VLA substantially accelerates inference while changing the number of successful trials by at most one across all combinations of the two models and two tasks. OpenVLA-OFT achieves only six successes out of 30 trials on FoldCloth, indicating limited baseline performance. TauPrune-VLA, which uses perceptual signals to guide token pruning, largely preserves this performance but does not yield an improvement on this task. For $\pi_{0.5}$, the numbers of successful trials are identical with and without TauPrune-VLA on both tasks. As with OpenVLA-OFT, inference is substantially accelerated, enabling more continuous execution on the physical robot.

On the TIANJI Marvin M6, TauPrune-VLA improves inference speed but slightly reduces the success rate. We attribute this decline to errors in grasp-point selection after pruning, given the challenges posed by the cable's slender shape and deformability. Overall, these results strongly support the ability of TauPrune-VLA to generalize across the evaluated VLA models and robotic platforms, while indicating room for further improvement on challenging tasks.

\begin{table}[t]
  \centering
  \caption{Ablation study of TauPrune-VLA.}
  \label{tab:tauprune_ablation}
  \small
  \setlength{\tabcolsep}{1pt}
  \setlength{\aboverulesep}{0pt}
  \setlength{\belowrulesep}{0pt}
  \renewcommand{\arraystretch}{1.18}
  \newcommand{\AblHead}[1]{{\footnotesize #1}}

  \begin{tabularx}{\columnwidth}{
    @{}l|*{5}{>{\centering\arraybackslash}X}|cc
  }
    \toprule
    \multirow{2}{*}{\AblHead{Method}}
      & \multicolumn{5}{c|}{\AblHead{Success Rate (\%) $\uparrow$}}
      & \AblHead{Latency} & \AblHead{Speedup} \\
    \cline{2-6}
      & \AblHead{Spatial} & \AblHead{Object} & \AblHead{Goal}
      & \AblHead{Long} & \AblHead{Avg.}
      & \AblHead{(ms) $\downarrow$} & \AblHead{$\uparrow$} \\
    \midrule
    Vanilla
      & 98.2 & 97.8 & 96.6 & \textbf{94.8} & 96.85 & 122.57 & 1$\times$ \\
    \midrule
    \multicolumn{8}{c}{\textit{Fixed $\tau$}} \\
    \midrule
    $\tau=0.001$
      & 98.4 & 96.4 & 97.6 & 90.2 & 95.65 & \textbf{65.38} & \textbf{1.87}$\times$ \\
    $\tau=0.25$
      & 98.2 & \textbf{98.2} & 97.2 & 93.2 & 96.70 & 65.95 & 1.86$\times$ \\
    $\tau=1$
      & 97.6 & 97.8 & 95.6 & 93.0 & 96.00 & 65.48 & \textbf{1.87}$\times$ \\
    \midrule
    \multicolumn{8}{c}{\textit{Architecture}} \\
    \midrule
    Async.\ MMTok
      & 97.8 & 97.2 & 95.4 & 92.8 & 95.80 & 66.01 & 1.86$\times$ \\
    Synchronous
      & \textbf{98.6} & \textbf{98.2} & \textbf{98.0} & 94.4 & \textbf{97.30}
      & 81.55 & 1.50$\times$ \\
    \midrule
    TauPrune-VLA
      & 98.2 & 97.6 & \textbf{98.0} & 94.6 & 97.10 & 65.82 & 1.86$\times$ \\
    \bottomrule
  \end{tabularx}
\end{table}

\subsection{Analysis}
\label{sec:analysis}

\textbf{Ablation Study.}
We evaluate two components of TauPrune-VLA, Perception--Action Temperature Modulation and Pipelined Token Pruning, using OpenVLA-OFT on LIBERO benchmark. The results are reported in Table~\ref{tab:tauprune_ablation}.

We first replace Perception--Action Temperature Modulation with a fixed temperature $\tau$ and evaluate three temperature settings. These replacements have a negligible effect on inference speed but slightly reduce the average success rate, supporting the effectiveness of temperature modulation. Similar to the results in Table~1, the fixed-temperature variants exhibit different trade-offs across task attributes, preserving performance on some attributes at the expense of others. This pattern further highlights the importance of an appropriate scale of perceptual representation for preserving VLA generalization across task attributes. The full TauPrune-VLA method achieves a higher average success rate than all evaluated fixed-$\tau$ variants, although some of these variants perform better on individual attributes. These results suggest that Perception--Action Temperature Modulation helps accommodate changes in the perceptual representation scale required for VLA decision-making, while leaving room for further improvement.

We further compare serial and asynchronous pruning architectures to assess the benefits of Pipelined Token Pruning. The serial variant of TauPrune-VLA computes pruning indices from the current frame and achieves an average success rate of 97.3\% at an inference latency of 81.55~ms. The asynchronous variant computes pruning indices in advance, reducing latency to 65.82~ms with a decrease of only 0.2\% in average success rate relative to the serial variant. To further examine the temporal stability of semantic structure in perceptual representations, we also apply asynchronous execution to MMTok, which likewise uses perceptual signals for token selection. This modification boosts inference speed by 10.18\%, with only a 0.15\% drop in average success rate, reducing latency at minimal performance cost.

\begin{figure}[t]
    \centering
    \includegraphics[width=0.9\columnwidth]{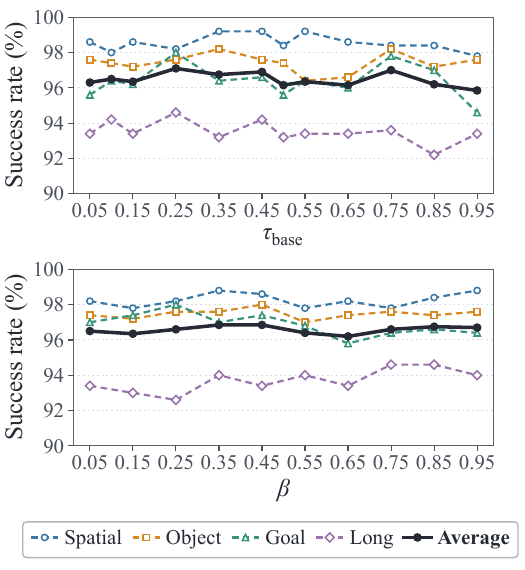}
    \caption{Impact on Hyperparameters $\tau_{base}$ and $\beta$.}
    \label{fig:chao}
\end{figure}

\textbf{Hyperparameter Sensitivity Analysis.} TauPrune-VLA exhibits only minor performance variations across the evaluated settings of $\tau_{\mathrm{base}}$ and $\beta$ on LIBERO (Fig.~\ref{fig:chao}). These hyperparameters regulate $\tau$ to guard against temperature collapse and uncontrolled growth, and the observed performance stability is consistent with effective temperature regulation. Individual suites exhibit different preferences for $\tau_{\mathrm{base}}$, most commonly favoring $0.25$ or $0.75$. We attribute these preferences to two effects of these settings: they allow $\tau$ to vary more broadly between $0$ and $1$ to accommodate changes in perceptual representation scale, and they limit the adverse effects of inappropriate temperature corrections.

\begin{figure}[t]
    \centering
    \includegraphics[width=0.9\columnwidth]{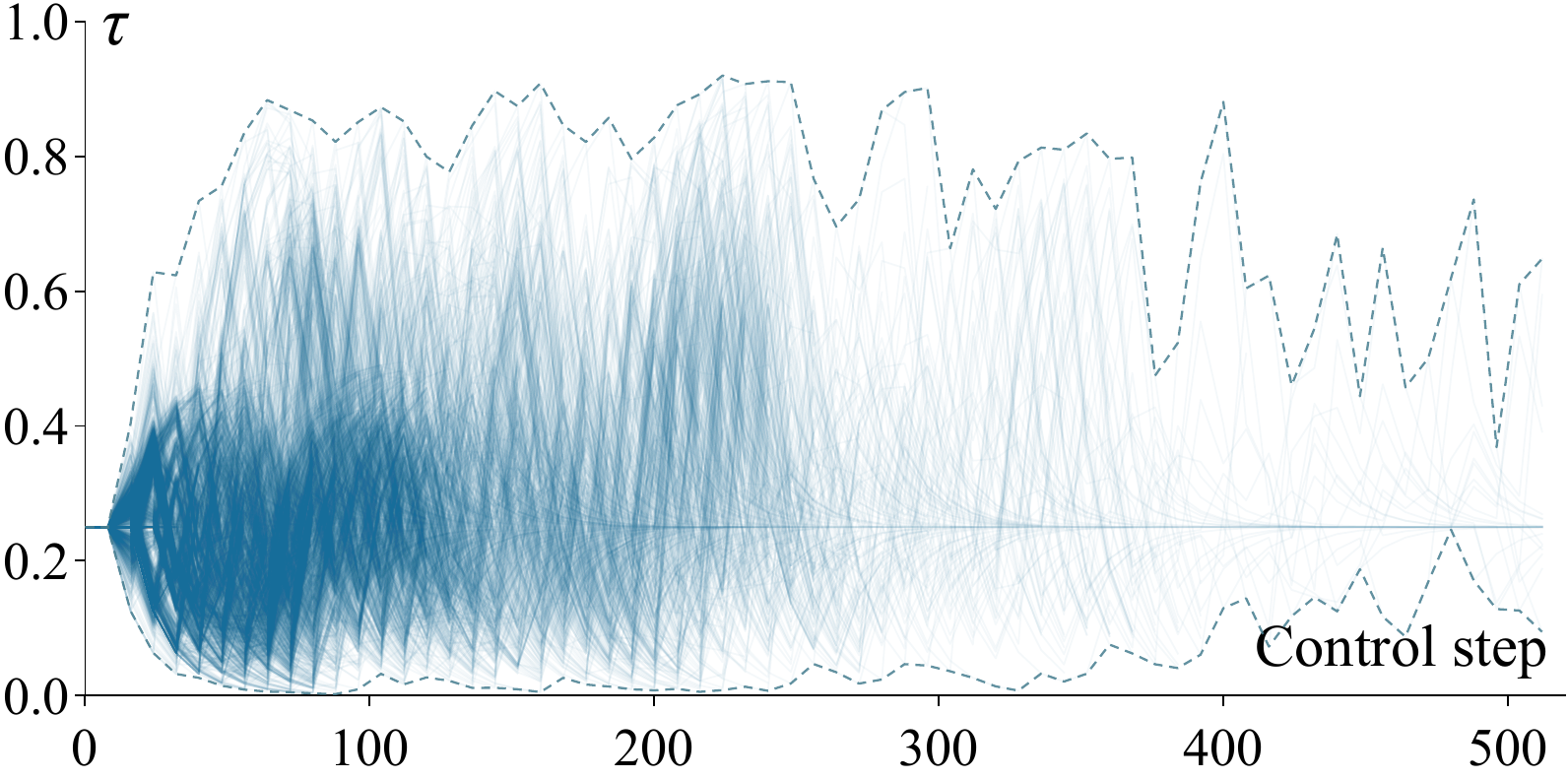}
    \caption{Trajectories of the perception temperature $\tau$.}
    \label{fig:tau}
\end{figure}

\textbf{Further Analysis of $\tau$.} Fig.~\ref{fig:tau} shows the evolution of $\tau$ across all LIBERO evaluation episodes, illustrating the behavior of Perception--Action Temperature Modulation over time. With $\tau_{\mathrm{base}}=0.25$, $\tau$ varies over a broad range, consistent with adaptation to changes in perceptual representation scale. Meanwhile, $\tau_{\mathrm{base}}$ provides a reference toward which excessively low or high temperatures are pulled, unless adjustments persist in the same direction. This behavior helps limit excessive temperature drift while allowing sustained and stable directional adjustments.

\section{Conclusion}
\label{sec:conclusion}

We presented TauPrune-VLA, a training-free framework for accelerating VLA inference through scale-adaptive perceptual token pruning. Our results support the finding that perceptual semantics can guide effective token selection without relying on internal attention scores when interpreted at a scale suited to the evolving scene structure. TauPrune-VLA controls the trade-off between cross-cluster coverage and within-cluster fidelity through a perception temperature, updated online using perceptual trends and modulated by robot-state changes and action magnitude. An asynchronous pipeline reuses retention indices across consecutive policy invocations to reduce pruning overhead. Evaluations across three VLA models, two simulation benchmarks, and two real-world robotic platforms demonstrate substantial acceleration while largely preserving task success, achieving up to a $1.86\times$ inference speedup.

\clearpage
\begingroup
\small
\bibliographystyle{plainnat}
\bibliography{main}

\begin{thebibliography}{30}
\providecommand{\natexlab}[1]{#1}
\providecommand{\url}[1]{\texttt{#1}}
\expandafter\ifx\csname urlstyle\endcsname\relax
  \providecommand{\doi}[1]{doi: #1}\else
  \providecommand{\doi}{doi: \begingroup \urlstyle{rm}\Url}\fi

\bibitem[Alvar et~al.(2025)Alvar, Singh, Akbari, and Zhang]{alvar2025divprune}
Saeed~Ranjbar Alvar, Gursimran Singh, Mohammad Akbari, and Yong Zhang.
\newblock Divprune: Diversity-based visual token pruning for large multimodal
  models.
\newblock In \emph{Proceedings of the IEEE/CVF Conference on Computer Vision
  and Pattern Recognition}, pages 9392--9401, 2025.

\bibitem[Brohan et~al.(2022)Brohan, Brown, Carbajal, Chebotar, and
  et~al]{brohan2022rt}
Anthony Brohan, Noah Brown, Justice Carbajal, Yevgen Chebotar, and et~al.
\newblock Rt-1: Robotics transformer for real-world control at scale.
\newblock \emph{arXiv preprint arXiv:2212.06817}, 2022.

\bibitem[Chen et~al.(2024)Chen, Zhao, Liu, Bai, and et~al]{chen2024image}
Liang Chen, Haozhe Zhao, Tianyu Liu, Shuai Bai, and et~al.
\newblock An image is worth 1/2 tokens after layer 2: Plug-and-play inference
  acceleration for large vision-language models.
\newblock In \emph{European Conference on Computer Vision}, pages 19--35, 2024.

\bibitem[Dong et~al.(2026)Dong, Hu, Zhang, Yin, Fu, and Qian]{dong2026mmtok}
Sixun Dong, Juhua Hu, Mian Zhang, Ming Yin, Yanjie Fu, and Qi~Qian.
\newblock Mmtok: Multimodal coverage maximization for efficient inference of
  vlms.
\newblock In \emph{Proceedings of International Conference on Learning
  Representations}, volume 2026, pages 48075--48099, 2026.

\bibitem[Fu et~al.(2024)Fu, Zhao, and Finn]{fu2024mobile}
Zipeng Fu, Tony~Z Zhao, and Chelsea Finn.
\newblock Mobile aloha: Learning bimanual mobile manipulation with low-cost
  whole-body teleoperation.
\newblock \emph{arXiv preprint arXiv:2401.02117}, 2024.

\bibitem[Huang et~al.(2026)Huang, Ding, Tang, Zhu, and
  et~al]{huang2026environment}
Yuting Huang, Leilei Ding, Zhipeng Tang, Zenghuan Zhu, and et~al.
\newblock Environment-aware adaptive pruning with interleaved inference
  orchestration for vision-language-action models.
\newblock In \emph{Proceedings of the International Conference on Machine
  Learning}, 2026.

\bibitem[Intelligence et~al.(2025)Intelligence, Black, Brown, Darpinian, and
  et~al]{intelligence2025pi_}
Physical Intelligence, Kevin Black, Noah Brown, James Darpinian, and et~al.
\newblock \(\backslash\pi\_ \{0.5\}\): a vision-language-action model with
  open-world generalization.
\newblock \emph{arXiv preprint arXiv:2504.16054}, 2025.

\bibitem[Karamcheti et~al.(2024)Karamcheti, Nair, Balakrishna, Liang, Kollar,
  and Sadigh]{karamcheti2024prismatic}
Siddharth Karamcheti, Suraj Nair, Ashwin Balakrishna, Percy Liang, Thomas
  Kollar, and Dorsa Sadigh.
\newblock Prismatic vlms: Investigating the design space of
  visually-conditioned language models.
\newblock \emph{arXiv preprint arXiv:2402.07865}, 2024.

\bibitem[Kim et~al.(2025{\natexlab{a}})Kim, Finn, and Liang]{kim2025fine}
Moo~Jin Kim, Chelsea Finn, and Percy Liang.
\newblock Fine-tuning vision-language-action models: Optimizing speed and
  success.
\newblock \emph{arXiv preprint arXiv:2502.19645}, 2025{\natexlab{a}}.

\bibitem[Kim et~al.(2025{\natexlab{b}})Kim, Pertsch, Karamcheti, Xiao, and
  et~al]{openvla}
Moo~Jin Kim, Karl Pertsch, Siddharth Karamcheti, Ted Xiao, and et~al.
\newblock {OpenVLA: An Open-Source Vision-Language-Action Model}.
\newblock In \emph{Proceedings of The Conference on Robot Learning}, volume
  270, pages 2679--2713, 2025{\natexlab{b}}.

\bibitem[Li et~al.(2024{\natexlab{a}})Li, Liang, Wang, Luo, and
  et~al]{li2024cogact}
Qixiu Li, Yaobo Liang, Zeyu Wang, Lin Luo, and et~al.
\newblock Cogact: A foundational vision-language-action model for synergizing
  cognition and action in robotic manipulation.
\newblock \emph{arXiv preprint arXiv:2411.19650}, 2024{\natexlab{a}}.

\bibitem[Li et~al.(2024{\natexlab{b}})Li, Hsu, Gu, Pertsch, and
  et~al]{li24simpler}
Xuanlin Li, Kyle Hsu, Jiayuan Gu, Karl Pertsch, and et~al.
\newblock Evaluating real-world robot manipulation policies in simulation.
\newblock \emph{arXiv preprint arXiv:2405.05941}, 2024{\natexlab{b}}.

\bibitem[Li et~al.(2026)Li, Meng, Sun, Ji, and et~al]{li2026spvla}
Ye~Li, Yuan Meng, Zewen Sun, Kangye Ji, and et~al.
\newblock {SP-VLA: A Joint Model Scheduling and Token Pruning Approach for VLA
  Model Acceleration}.
\newblock In \emph{Proceedings of International Conference on Learning
  Representations}, pages 38155--38184, 2026.

\bibitem[Lin et~al.(2026)Lin, Zhong, Du, Zhang, and et~al]{lin2026evo}
Tao Lin, Yilei Zhong, Yuxin Du, Jingjing Zhang, and et~al.
\newblock Evo-1: Lightweight vision-language-action model with preserved
  semantic alignment.
\newblock In \emph{Proceedings of IEEE/CVF Conference on Computer Vision and
  Pattern Recognition}, pages 13397--13406, 2026.

\bibitem[Liu et~al.(2023{\natexlab{a}})Liu, Zhu, Gao, Feng, and
  et~al]{liu2023libero}
Bo~Liu, Yifeng Zhu, Chongkai Gao, Yihao Feng, and et~al.
\newblock Libero: Benchmarking knowledge transfer for lifelong robot learning.
\newblock \emph{Proceedings of Advances in Neural Information Processing
  Systems}, 36:\penalty0 44776--44791, 2023{\natexlab{a}}.

\bibitem[Liu et~al.(2023{\natexlab{b}})Liu, Li, Wu, and Lee]{llava}
Haotian Liu, Chunyuan Li, Qingyang Wu, and Yong~Jae Lee.
\newblock {Visual Instruction Tuning}.
\newblock In \emph{Proceedings of Advances in Neural Information Processing
  Systems}, volume~36, pages 34892--34916, 2023{\natexlab{b}}.

\bibitem[Liu et~al.(2024)Liu, Liu, Wang, An, and et~al]{liu2024robomamba}
Jiaming Liu, Mengzhen Liu, Zhenyu Wang, Pengju An, and et~al.
\newblock {RoboMamba: Efficient Vision-Language-Action Model for Robotic
  Reasoning and Manipulation}.
\newblock In \emph{Proceedings of Advances in Neural Information Processing
  Systems}, volume~37, pages 40085--40110, 2024.
\newblock \doi{10.52202/079017-1266}.

\bibitem[Liu et~al.(2025)Liu, Chen, Cai, Lin, and et~al]{liu2025vla}
Ziyan Liu, Yeqiu Chen, Hongyi Cai, Tao Lin, and et~al.
\newblock Vla-pruner: Temporal-aware dual-level visual token pruning for
  efficient vision-language-action inference.
\newblock \emph{arXiv e-prints}, pages arXiv--2511, 2025.

\bibitem[Ma et~al.(2026)Ma, Zhang, Wang, Wang, and et~al]{ma2026safe}
Shilin Ma, Chubin Zhang, Changyuan Wang, Yuji Wang, and et~al.
\newblock Safe-pruner: Semantic attention--guided future-aware token pruning
  for efficient vision-language-action manipulation.
\newblock In \emph{Proceedings of European Conference on Computer Vision},
  pages 163--180, 2026.

\bibitem[Pei et~al.(2026)Pei, Chen, Xu, Wang, and et~al]{pei2026action}
Xiaohuan Pei, Yuxing Chen, Siyu Xu, Yunke Wang, and et~al.
\newblock Action-aware dynamic pruning for efficient vision-language-action
  manipulation.
\newblock In \emph{Proceedings of International Conference on Learning
  Representations}, volume 2026, pages 10832--10851, 2026.

\bibitem[Shukor et~al.(2025)Shukor, Aubakirova, Capuano, Kooijmans, and
  et~al]{shukor2025smolvla}
Mustafa Shukor, Dana Aubakirova, Francesco Capuano, Pepijn Kooijmans, and
  et~al.
\newblock {SmolVLA: A Vision-Language-Action Model for Affordable and Efficient
  Robotics}.
\newblock arXiv:2506.01844, 2025.
\newblock URL \url{https://arxiv.org/abs/2506.01844}.

\bibitem[Sun et~al.(2026)Sun, Xia, Li, Ren, Cui, and et~al]{sun2026makes}
Luoyang Sun, Guoyang Xia, Fengfa Li, Lei Ren, Xinyu Cui, and et~al.
\newblock What makes an efficient vla? navigating action-head design, scaling,
  and latency.
\newblock \emph{arXiv preprint arXiv:2609.13984}, 2026.

\bibitem[Team et~al.(2024)Team, Riviere, Pathak, Sessa, and
  et~al]{team2024gemma}
Gemma Team, Morgane Riviere, Shreya Pathak, Pier~Giuseppe Sessa, and et~al.
\newblock Gemma 2: Improving open language models at a practical size.
\newblock \emph{arXiv preprint arXiv:2408.00118}, 2024.

\bibitem[Touvron et~al.(2023)Touvron, Martin, Stone, Albert, and
  et~al]{touvron2023llama}
Hugo Touvron, Louis Martin, Kevin Stone, Peter Albert, and et~al.
\newblock Llama 2: Open foundation and fine-tuned chat models.
\newblock \emph{arXiv preprint arXiv:2307.09288}, 2023.

\bibitem[Wang et~al.(2026)Wang, Xu, Xiang, Pan, and et~al]{wang2026specprune}
Hanzhen Wang, Jiaming Xu, Yushun Xiang, Jiayi Pan, and et~al.
\newblock Specprune-vla: Accelerating vision-language-action models via
  action-aware self-speculative pruning.
\newblock In \emph{Proceedings of International Conference on Machine
  Learning}, 2026.

\bibitem[Wei et~al.(2026)Wei, Fan, Guo, Zhen, and et~al]{wei2026learning}
Yujie Wei, Jiahan Fan, Jiyu Guo, Ruichen Zhen, and et~al.
\newblock Learning to accelerate vision-language-action models through adaptive
  visual token caching.
\newblock \emph{arXiv preprint arXiv:2602.00686}, 2026.

\bibitem[Xu et~al.(2026)Xu, Wang, Xia, Zhu, and et~al]{xu2026vla}
Siyu Xu, Yunke Wang, Chenghao Xia, Dihao Zhu, and et~al.
\newblock Vla-cache: Efficient vision-language-action manipulation via adaptive
  token caching.
\newblock \emph{Proceedings of Advances in Neural Information Processing
  Systems}, 38:\penalty0 164448--164473, 2026.

\bibitem[Yang et~al.(2025)Yang, Wang, Wen, Zhongwei, and
  et~al]{yang2025efficientvla}
Yantai Yang, Yuhao Wang, Zichen Wen, Luo Zhongwei, and et~al.
\newblock {EfficientVLA: Training-Free Acceleration and Compression for
  Vision-Language-Action Models}.
\newblock In \emph{Proceedings of Advances in Neural Information Processing
  Systems}, volume~38, pages 40891--40914, 2025.

\bibitem[Zhang et~al.(2026)Zhang, Liu, Li, Lu, and et~al]{zhang2026beyond}
Qizhe Zhang, Mengzhen Liu, Lichen Li, Ming Lu, and et~al.
\newblock Beyond attention or similarity: Maximizing conditional diversity for
  token pruning in mllms.
\newblock \emph{Proceedings of Advances in Neural Information Processing
  Systems}, 38:\penalty0 25438--25468, 2026.

\bibitem[Zitkovich et~al.(2023)Zitkovich, Yu, Xu, Xu, and et~al]{rt2}
Brianna Zitkovich, Tianhe Yu, Sichun Xu, Peng Xu, and et~al.
\newblock {RT-2: Vision-Language-Action Models Transfer Web Knowledge to
  Robotic Control}.
\newblock In \emph{Proceedings of The Conference on Robot Learning}, volume
  229, pages 2165--2183, 2023.

\end{thebibliography}
\endgroup

\end{document}